\documentclass[10pt,journal,compsoc]{IEEEtran}

\usepackage{graphicx}
\usepackage{comment}
\usepackage{amsmath,amssymb} 
\usepackage{color}
\usepackage{extarrows}
\usepackage{pgfmath}

\usepackage{bm}
\usepackage{subcaption}
\usepackage[flushleft]{threeparttable}
\usepackage{tablefootnote}
\usepackage{booktabs} 
\usepackage{tcolorbox}
\usepackage{algorithm}
\usepackage{algpseudocode}
\usepackage[colorlinks=true,linkcolor=black,anchorcolor=black,citecolor=black,filecolor=black,menucolor=black,runcolor=black,urlcolor=black]{hyperref}
\usepackage{multirow}
\usepackage{diagbox}
\usepackage{standalone}

\def\ourmethod{\emph{HERS}}
\newcommand{\etal}{et al.}
\newcommand{\floor}[1]{\left\lfloor #1 \right\rfloor}
\newcommand{\ceil}[1]{\left\lceil #1 \right\rceil}
\newcommand{\round}[1]{\left\lfloor #1 \right\rceil}
\newcommand{\modulo}[1]{[ #1 ]}
\newcommand{\specialcell}[2][c]{\begin{tabular}[#1]{@{}c@{}}#2\end{tabular}}

\DeclareMathOperator*{\argmax}{arg\,max}

\ifCLASSOPTIONcompsoc
  \usepackage[nocompress]{cite}
\else
  \usepackage{cite}
\fi

\begin{document}

\title{HERS: Homomorphically Encrypted Representation Search}

\author{Joshua~J.~Engelsma,~\IEEEmembership{Student Member,~IEEE,}
        Anil~K.~Jain,~\IEEEmembership{Life Fellow,~IEEE,}
        and~Vishnu~Naresh~Boddeti,~\IEEEmembership{Member,~IEEE}
\IEEEcompsocitemizethanks{\IEEEcompsocthanksitem Authors are with the Department of Computer Science and Engineering, Michigan State University, East Lansing, MI, 48824.\protect\\
E-mail: \{engelsm7, jain, vishnu\}@msu.edu
}
}

\markboth{IEEE Transactions on Biometrics, Behavior, and Identity Science}%
{Engelsma \MakeLowercase{\textit{et al.}}: Homomorphically Encrypted Representation Search}

\IEEEtitleabstractindextext{%
\begin{abstract}
We present a method to search for a probe (or query) image representation against a large gallery in the encrypted domain. We require that the probe and gallery images be represented in terms of a fixed-length representation, which is typical for representations obtained from learned networks. Our encryption scheme is agnostic to how the fixed-length representation is obtained and can therefore be applied to any fixed-length representation in any application domain. Our method, dubbed \ourmethod{}~(Homomorphically Encrypted Representation Search), operates by (i) compressing the representation towards its estimated intrinsic dimensionality with \textit{minimal loss} of accuracy (ii) encrypting the compressed representation using the proposed fully homomorphic encryption scheme, and (iii) efficiently searching against a gallery of encrypted representations \textit{directly in the encrypted domain, without decrypting them}. Numerical results on large galleries of face, fingerprint, and object datasets such as ImageNet show that, for the first time, accurate and fast image search within the encrypted domain is feasible at scale (500 seconds; $275\times$ speed up over state-of-the-art for encrypted search against a gallery of 100 million). Code is available at \url{https://github.com/human-analysis/hers-encrypted-image-search}
\end{abstract}
}

\maketitle
\IEEEdisplaynontitleabstractindextext
\IEEEpeerreviewmaketitle

\IEEEraisesectionheading{\section{Introduction}\label{sec:introduction}}

\IEEEPARstart{I}{n} 2014, a hack on the US Office of Personnel Management (OPM) left 22 million user records exposed, including millions of fingerprint records. Data breaches, like the OPM hack, could have untold consequences against those whose personal identifiable information (PII~\footnote{\tiny\url{https://bit.ly/2HD83Pq}}) was compromised, including identity theft, robbery, unauthorized access to secure facilities, and blackmail. Sadly, in today's day and age, data-breaches like these are not isolated incidents~\footnote{\tiny \url{https://wapo.st/39PQuaT}}~\footnote{\tiny\url{https://wapo.st/2V3kHPS}}~\footnote{\tiny\url{https://bit.ly/2OQhlM3}}, motivating and necessitating the development of strong encryption techniques which protect the underlying data \textit{at all times}.

\begin{figure*}[t]
\begin{center}
\includegraphics[width=\textwidth]{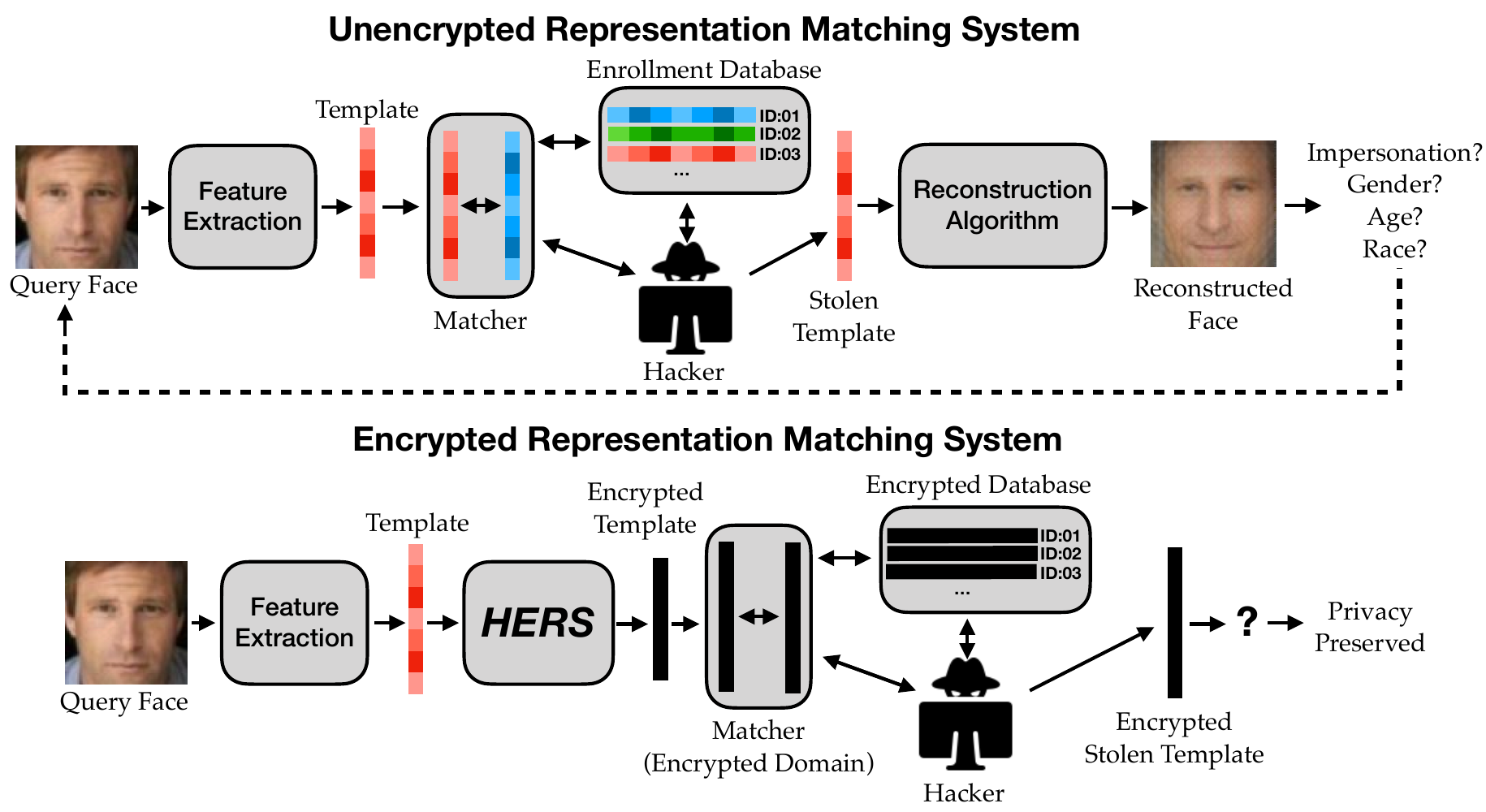}
\caption{\textbf{Overview:} Unencrypted and Encrypted face image search systems. Given an image representation (template), it is either (i) stored in a database (enrollment) or (ii) passed to the matcher for searching. Both the database and the matcher are potential points of attack. An attacker can either steal a template directly from an unencrypted database, launching an impersonation attack. Or, if the database were AES encrypted, the hacker could attack the matcher where templates are decrypted for comparison. In our approach, (i) the templates are encrypted during database enrollment and (ii) the templates are matched \textit{within the encrypted domain}. As such, a hacker is unable to exploit any meaningful information from stolen templates.\label{fig:intro_fig}}
\end{center}
\end{figure*}

Perhaps the most vulnerable category of stored data still needing adequate protection is that of image representations (\textit{e.g.} face representations). While many forms of data can be sufficiently secured in a database with a well-known encryption scheme like the Advanced Encryption Standard (AES)~\cite{nist_aes}, image representations present a unique challenge. In particular, query image representations are often searched against other representations already enrolled in the database (\textit{e.g.} a face search system). Even if the enrolled representations were protected with AES, they would have to be decrypted prior to matching with a query representation, leaving both of them vulnerable at the point of comparison~(Fig.~\ref{fig:intro_fig}). As such, it is critically important to develop strong encryption techniques to protect image representations in the database and during matching.

While much research continues in the area of improving the discriminative power of image representations (\textit{e.g.} face recognition~\cite{face1,facenet} and image classification~\cite{russakovsky2015imagenet}), relatively little effort has been invested into ensuring the security of the representations after they have been learned. This is alarming considering deep face representations can be (i) reconstructed back into their corresponding face image (Fig.~\ref{fig:intro_fig})~\cite{mai2018reconstruction} or (ii) violate user privacy by mining demographic attributes such as age, ethnicity, or gender~\cite{liu2015deep}. More generally, it is well known that local features, such as deep CNN representations~\cite{dosovitskiy2016inverting}, SIFT~\cite{weinzaepfel2011reconstructing,pittaluga2019revealing}, HOG~\cite{vondrick2013hoggles} and Bag-of-Visual-Words~\cite{kato2014image}, can be inverted back into the image space with high fidelity.

A special class of encryption algorithms which enable basic arithmetic operations (multiplications and/or additions) in the encrypted domain are known as homomorphic encryption (HE) systems~\cite{fhe3}. Since representations extracted by CNNs are typically compared using distance metrics that can be expressed in terms of additions and multiplications, like euclidean distance or cosine similarity, HE systems are a plausible solution to protect the representations within the database, and also during matching. HE systems are an attractive solution for secure feature matching systems. They have many desirable properties~\cite{maltoni2009handbook} including  \textbf{diversity}, \textbf{revocability}, \textbf{security}, \textbf{performance}, and \textbf{privacy}. At the same time they do not suffer from any loss in matching performance since the encrypted computations are exact. Figure~\ref{fig:intro_fig} shows a comparison of template matching in the unencrypted and the encrypted domain.

The key barrier against using HE schemes ``off-the-shelf"~\cite{fhe2} for protection of image representations is the computational complexity of arithmetic operations directly on the ciphertext (in the encrypted domain). This is especially true of \textit{fully}-homomorphic encryption schemes (FHE) which enable both additions \textit{and} multiplications within the encrypted domain~\cite{fhe2}. For example, the authors in~\cite{troncoso2013fully} showed that a naive implementation of FHE requires 48.7 MB of memory and 12.8 seconds to match a single pair of 512-dimensional encrypted face representations. Such computational requirements restrict the application of FHE schemes in $1:1$ matching applications and completely render them impractical in $1:m$ search applications at scale.

To overcome these limitations, we present a practical solution, dubbed \ourmethod{}, for $1:m$ encrypted feature matching at scale. This is achieved through a synergistic combination of dimensionality reduction of features and development of a more efficient FHE scheme. The specific contributions of this paper are as follows,
\begin{enumerate}
  \item A data encoding scheme that is tailored for efficient $1:m$ representation matching in the encrypted domain by leveraging SIMD\footnote{Single Instruction Multiple Data} capabilities of existing FHE schemes. Over a gallery of 1 Million 512-dim templates, this provides an $11\times$ speed-up.
  \item A dimensionality reduction scheme, dubbed DeepMDS++, based on DeepMDS~\cite{gong2019intrinsic}, a state-of-the-art dimensionality reduction technique. For a 192-dim fingerprint representation from DeepPrint~\cite{deepprint}, this provides a 6$\times$ speed-up at an only $0.1\%$ loss in rank-1 search accuracy against a gallery of 100 million.
  \item Extensive experimental analysis (face, fingerprint, and ImageNet datasets) in terms of accuracy, latency and memory requirements when performing image search in the encrypted domain. Our results indicate that \ourmethod{}~is the first scheme capable of delivering accurate (within $\approx$ 2\% of unencrypted accuracy) and timely (within 10 minutes) image search in the encrypted domain at scale (100 million gallery). The overall scheme provides a $186\times$ speed-up over a state-of-the-art $1:1$ matching of 32-dim encrypted feature vectors.
\end{enumerate}

\begin{table}[t]
\caption{Search Speed: 1 Million Gallery}
 \centering
\begin{threeparttable} 
\begin{tabular}{ cc }
\toprule
 \specialcell{Baseline Speed\tnote{1}} & \specialcell{Baseline Processor Speed\tnote{\textdagger}}\\
\midrule
~\cite{rw1}~13,000 seconds\tnote{2} & 2.4 GHz\\
\midrule
~\cite{boddeti2018secure}~ 10,000 seconds & 3.3 GHz\\
\midrule
~\cite{rw2}~ 12.7 Million seconds & 3.3 GHz\\
\midrule
~\cite{rw3}~ 500 seconds\tnote{3} & 2.67 GHz\\
\midrule
~\cite{rw4}~ 850,000 seconds & 2.5 GHz\\
\midrule
~\cite{rw5}~ 109,000 seconds & 3.4 GHz\\
\midrule
~\cite{rw6}~ 2 Million seconds & 3.4 GHz\\
\bottomrule
\end{tabular}
\begin{tablenotes}
\item [\textdagger] HERS benchmarked on 3.3 GHz processor.
\item [1] HERS search speed against 1 million via 128-D templates is 300 seconds.
\item [2] Only 12-D features are used in~\cite{rw1}.
\item [3] ~\cite{rw3} uses Paillier Homomorphic encryption. It requires the client template to remain decrypted. HERS encrypts both client and gallery template at all times using FHE.
\end{tablenotes}
\end{threeparttable}
\label{table:comparison}
\end{table}

\section{Related Work}

\noindent\textbf{Privacy-Preserving Biometric Representations:} Methods to secure representations of personally identifying information have been developed over the past decade~\cite{rathgeb2011survey}. These are summarized in~\cite{boddeti2018secure} into two categories: (i) cryptographic protection, and (ii) pattern recognition based protection. Visual cryptography~\cite{visual1} is a common cryptographic approach for securing biometric data such as fingerprint~\cite{encrypt2} and face~\cite{encrypt4} images. Under some schemes, the visual perturbations need to be removed prior to performing a match, exposing the biometric data during authentication or search. Fuzzy-vaults~\cite{fuzzy1} are another cryptosystem which have been utilized for fingerprint~\cite{fuzzy3} and iris~\cite{fuzzy5} recognition. Pattern recognition based protection schemes have been proposed as an alternative to cryptosystems. Examples include non-invertible transformation functions~\cite{pr2} and cancelable biometrics~\cite{pr3, jin2017ranking}. Additionally, key-binding systems have been proposed to merge a biometric template with a secret key~\cite{pr4,pr5}. More recently, template security approaches based on neural networks \cite{mai2020secureface} and representation geometry \cite{kim2021ironmask} have been designed for face representations. All of these approaches trade-off matching performance for security of the template. In contrast, \ourmethod~does not suffer from this trade-off i.e., it provides very high levels of representation security with minimal loss in matching accuracy.

\vspace{5pt}
\noindent\textbf{Privacy-Preserving Visual Recognition:} Within the broader context of computer vision there is growing interest in privacy-preserving techniques. These approaches are based on cryptographic methods or computer vision techniques. Cryptographic methods include face detection~\cite{avidan2006blind,avidan2007efficient} using secure multi-party computation, image retrieval~\cite{shashank2008private} by oblivious transfer (a building block of secure multi-party computation), face verification~\cite{upmanyu2010blind} using the Paillier Cryptosystem, video surveillance~\cite{upmanyu2009efficient} using Secret Sharing, learning or inference over private data via differential privacy~\cite{abadi2016deep} and homomorphic encryption~\cite{gilad2016cryptonets,yonetani2017privacy, brutzkus2019low, nalini1, nalini2, nalini3}. Computer vision techniques include camera localization~\cite{speciale2019privacy1,speciale2019privacy2} by lifting 2D and 3D points to 2D and 3D lines, embedding features into an adversarially designed subspace \cite{dusmanu2020privacy}, detecting private computer screens~\cite{korayem2016enhancing} using CNNs, and activity recognition~\cite{ren2018learning,ryoo2017privacy} through image manipulation. In contrast to the foregoing, \ourmethod~adopts a synergistic combination of computer vision techniques in the form of dimensionality reduction and cryptographic methods (FHE scheme), resulting in both efficient and accurate representation matching in the encrypted domain at scale.

\vspace{5pt}
\noindent\textbf{Encrypted Distance Computation:} Homomorphic encryption (HE) cryptosystems enable arithmetic directly on ciphertext, and as such can be leveraged to compute distances between feature vectors entirely in the encrypted domain. However, given the extreme computational complexity of HE~\cite{fhe3} most existing works utilizing HE are limited to binary templates and partial homomorphic encryption (PHE) (supports either encrypted additions or encrypted multiplications), or are computationally impractical~\cite{he3,he4, gomez2017multi, rw1, rw2, rw3, rw4, rw5, rw6, rw7, rw8}. From Table~\ref{table:comparison} it can be seen that existing methods are orders of magnitude slower than even the best existing baseline~\cite{boddeti2018secure}. The only exception to this is the work described in~\cite{rw3}, which uses Paillier Homomorphic encryption (requiring that the client template remain decrypted). Admittedly, many of the benchmarks reported in Table~\ref{table:comparison} are benchmarked on slower running CPUs than \ourmethod. However, given the order of magnitude in time difference between \ourmethod~and the benchmarks, it is not plausible that a minor CPU upgrade will come close to bridging the gap in search speed. A few works~\cite{he6, boddeti2018secure, kolberg2020efficiency} have demonstrated the effective use of Fully Homomorphic Encryption (FHE) schemes~\cite{fhe2} (supports encrypted multiplications and additions). Cheon \etal{} \cite{he6} and Boddeti \cite{boddeti2018secure} proposed FHE based schemes which leverage a batching technique to reduce the memory and computational requirements for $1:1$ matching of binary iris (Hamming distance) and quantized face representations (cosine distance), respectively, in the encrypted domain. Boddeti showed that two 512-dimensional face representations could be compared with 16 KB of memory and in 2.5 milliseconds. Engelsma \etal{} \cite{deepprint} adopted the same scheme for $1:1$ matching of 192-dimensional encrypted fingerprint representations in 1.25 milliseconds. In contrast, early FHE schemes for the same took 100 seconds per match~\cite{boddeti2018secure}.

Although these algorithmic savings enable using FHE for $1:1$ matching applications, the time and space complexity remains intractable for most $1:m$ matching applications (image search). For example, encrypted search against a $m=100,000,000$ gallery with representations that are of dimensionality $512$ would still take over 38 hours and 9 TB memory with the improved FHE scheme proposed in~\cite{boddeti2018secure}. As such, in this work we develop a FHE based solution that is explicitly designed for efficient $1:m$ encrypted matching and can be applied to any image search application.

\vspace{5pt}
\noindent{\textbf{Compressing Representations:}} A plethora of work has been done to estimate a low dimensional approximation of a data manifold which resides in a high dimensional space. In addition to the time tested classical methods for this task (\textit{e.g.,} PCA, MDS, or Isomap), methods utilizing deep networks to learn more complex non-linear mappings have also been proposed. Some of these approaches combine representation learning together with dimensionality reduction via a primary learning objective (\textit{e.g.} a classification loss) paired with a regularization loss term~\cite{dr1, dr2}. In contrast, other approaches~\cite{hinton2006reducing, hadsell2006dimensionality, gong2019intrinsic} train a completely separate dimensionality reduction algorithm on top of existing representations, such that access to the representation extraction network is not required. Since we want \ourmethod~to be extendable as a wrapper on top of any existing image search system, we pattern the dimensionality reduction component of \ourmethod{} after DeepMDS~\cite{gong2019intrinsic}, and significantly improve its performance.

\section{Approach}
\begin{figure*}[t]
\begin{center}
\includegraphics[width=\textwidth]{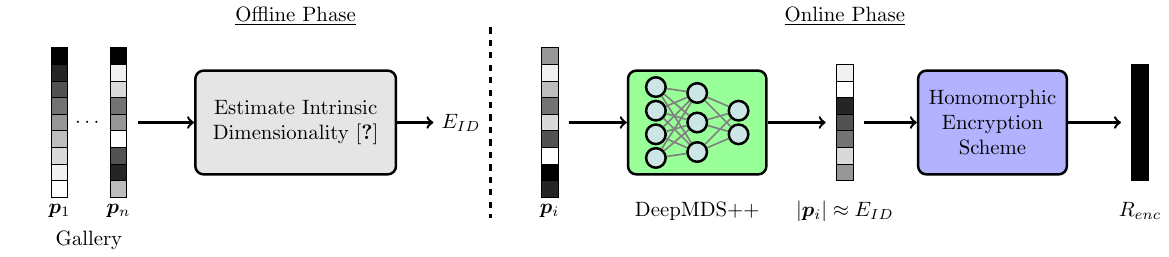}
\caption{Schematic Diagram of \ourmethod{}. First, we estimate the intrinsic dimensionality of a given representation~\cite{gong2019intrinsic} in an offline stage from a gallery $\bm{p}_1,\dots,\bm{p}_m$. Subsequently, we reduce the dimensionality of the representation towards its intrinsic dimensionality ($E_{ID}$) as much as possible, such that minimal accuracy is lost, using a deep network based non-linear mapping (DeepMDS++). Finally, the compressed representation $\bm{p}_i$ is homomorphically encrypted $R_{Enc}$ and passed on to our fast encrypted search algorithm.\label{fig:schematic_fig}}
\end{center}
\end{figure*}
We posit that there are two avenues for improving search efficiency in the encrypted domain: (i) the \textbf{encryption scheme} itself can be optimized for faster search, and (ii) the \textbf{dimensionality of the representation} can be compressed as far as possible, such that no accuracy is lost. We achieve fast and accurate search at scale in the encrypted domain by coupling both of these solutions in \ourmethod~(Fig.~\ref{fig:schematic_fig}). In the following sections, we elaborate on each of these steps individually.

\subsection{Problem Setup}
A typical representation matching task involves a database of $m$ template feature vectors $\bm{P}=[\bm{p}_1,\dots,\bm{p}_m]\in\mathbb{R}^{d\times m}$ against which a query representation $\bm{q}\in\mathbb{R}^{d}$ is matched. The result of the matching process is a score that determines the degree of similarity between $\bm{q}$ and each template in $\bm{P}$. A common metric that is adopted for template matching is the Euclidean distance or the cosine similarity. At the core, both of these metrics involve matrix-vector products of the form: $r(\bm{P},\bm{q}) = \bm{P}^T\bm{q}$. Hence, the representation matching process is comprised of $md$ scalar multiplications and $md$ scalar additions.

We devise a solution to cryptographically guarantee the security of the database $\bm{P}$ as well as the query $\bm{q}$ to prevent leakage of any private information. This can be achieved through a parameterized function that transforms a representation $\bm{z}$ from the original space into an alternate space, i.e., $\mathcal{E}(\bm{z}) = f(\bm{z};\bm{\theta}_{pk})$, where $f(\cdot;\bm{\theta}_{pk})$ is the encryption function with public-key $\bm{\theta}_{pk}$ and $\bm{z} = g(\mathcal{E}(\bm{z});\bm{\theta}_{sk})$, where $g(\cdot;\bm{\theta}_{sk})$ is the decryption function with private-key $\bm{\theta}_{sk}$. The key idea of our paper is to adopt an encryption function to secure the database and the query while retaining our ability to compute their matching score efficiently at scale and without any loss of accuracy i.e.,
\begin{equation}
  r(\bm{P},\bm{q}) \approx g\left(r(f(\bm{P},\bm{\theta}_{pk}),f(\bm{q},\bm{\theta}_{pk}));\bm{\theta}_{sk}\right)
\end{equation}

FHE satisfies this property and enables us to preserve user privacy. Even if a malicious attacker can gain access to the database of feature representations or the query, without access to the private-key $\bm{\theta}_{sk}$ the attacker cannot reconstruct the underlying image or extract any other information inherent to the representation.

\subsection{Fast and Secure Similarity Computation}
We use the Fan-Vercauteren (FV) scheme \cite{fan2012somewhat} as our base FHE scheme\footnote{A potentially more efficient but approximate alternative is CKKS~\cite{cheon2017homomorphic}, which can directly be applied to real-valued arithmetic.}. The mathematical basis of this scheme lies in modular arithmetic. Building upon this scheme, we propose a data encoding technique that is tailored for efficient $1:m$ representation matching. We now briefly describe the different components of our approach.

\vspace{5pt}
\noindent\textbf{Fan-Vercauteren Scheme:} Plaintext space of the FV scheme is represented as a polynomial ring over a finite field $R_t=\mathbb{Z}_t[x]/\bm{\Phi}_n(x)$, where $t\in\mathbb{Z}$ is an integer and $\bm{\Phi}_n(x)$ is an irreducible polynomial of degree at most $n-1$. Upon encrypting the plaintext polynomial, the encrypted numbers (ciphertext) are encoded as polynomials in the ring $R_q$. The FV scheme utilizes three keys, (1) a private decryption key $\bm{\theta}_{sk}$, (2) a public encryption key $\bm{\theta}_{pk}$, and (3) evaluation keys $\bm{\theta}_{ev}$ which are necessary for multiplication over encrypted data. Addition and multiplication of two ciphertexts translates to polynomial addition and dyadic multiplication in $R_q$. As long as the coefficients of the resulting polynomials do not exceed $q$, correctness is ensured. The exact description of the entire scheme, including encryption, decryption, ciphertext addition and ciphertext multiplication, is deferred to Appendix~\ref{sec:fv-scheme}.

\vspace{5pt}
\noindent\textbf{Encoding Scheme:} The FV scheme is designed to act on integers only. As such, we need to encode our real valued representation $\bm{q}\in\mathbb{R}^{d}$ into an integer valued representation $\bm{q}\in\mathbb{Z}^{d}$. We quantize a given representation's real-valued features into integers with a precision of $0.004$ and represent these integers in base $w$. The utility of the FV scheme is critically dependent on the encoding scheme chosen to represent the quantized features in the ring $R_t$. Therefore, to maintain the integrity of ciphertext computations, the choice of the ring $R_t$ needs to ensure that the range of values after the desired ciphertext operations remain within the same ring.

\begin{table*}[t]
    \centering
    \caption{Computational Complexity (\# of homomorphic operations) of matching a $d$ dimensional encrypted representation against an encrypted gallery of size $m$.\label{table:compute}}
    \begin{tabular}{l|cc|cc|c|c}
        \toprule
        Encoding Scheme & Multiplication & Ratio & Addition & Ratio & Rotation & Memory\\
        \midrule
        Na{\"i}ve & $md$ & 1 & $m(d-1)$ & 1 & 0 & $\mathcal{O}(mdn)$ \\
        Boddeti~\cite{boddeti2018secure} & $m$ & d & $m\log_2 d$ & $\frac{d-1}{\log_2 d}$ & $m\log_2 d$ & $\mathcal{O}(mn)$ \\
        \ourmethod (ours) & $\ceil{\frac{m}{n}}d$ & $\frac{m}{\ceil{\frac{m}{n}}}$ & $\ceil{\frac{m}{n}}(d-1)$ & $\frac{m}{\ceil{\frac{m}{n}}}$ & 0 & $\mathcal{O}\left(dn\ceil{\frac{m}{n}}\right)$ \\
        \bottomrule
    \end{tabular}
\end{table*}

Our key contribution in this paper is a custom encoding scheme for efficient $1:m$ matching by utilizing the SIMD primitives~\cite{juvekar2018gazelle} of the FV scheme. The primitives operate over an array of numbers instead of a single number, encoding multiple numbers within the same polynomial using the Chinese Remainder Theorem. The encoding scheme in Boddeti~\cite{boddeti2018secure} is also based on the same principles but is specifically suitable for $1:1$ matching.

Given a query $\bm{q}\in\mathbb{Z}^{d}$ and a database of feature vectors $\bm{P}\in\mathbb{Z}^{d\times m}$, the encoding scheme in \cite{boddeti2018secure} encodes the feature vector of each sample (column) into a single polynomial. In contrast, in \ourmethod, the client node encodes each dimension (row) of the representation into a separate polynomial. A query $\bm{q}$ is therefore encoded into $d$ plaintexts as,
\begin{equation}
\label{eq:query}
g_i = \sum_{j=1}^{m}\bm{q}[i]x^{j-1} \mbox{ } \forall i\in\{1,\dots,d\}
\end{equation}
\noindent Similarly, for the database, each dimension of all the templates $\bm{P}_{i\cdot}\in\mathbb{Z}^{m}$ are encoded into polynomial, resulting in $d$ plaintexts,
\begin{equation}
\label{eq:gallery}
h_i=\sum_{j=1}^{m}\bm{p}_j[i]x^{j-1} \mbox{ } \forall i\in\{1,\dots,d\}
\end{equation}
In this case, the polynomial which encodes the inner products of the query $\bm{q}$ and the templates $\bm{P}$ can be obtained as,
\begin{equation}
    s = \sum_{i=1}^d g_i * h_i = \sum_{j=1}^{n}<\bm{q},\bm{p}_j>x^{j-1}
\end{equation}
\noindent where the product $g * h$ is standard polynomial multiplication. As a consequence, the $m < n$ inner products can be computed through $d$ polynomial multiplications\footnote{When $m > n$ we can chunk the $m$ samples into $\ceil{\frac{m}{n}}$ databases of $n$ samples each.}. The FV encryption scheme, the corresponding cryptographic primitives, and ciphertext addition and multiplication, operate on this plaintext representation.

\vspace{5pt}
\noindent\textbf{Enrollment Protocol:} The client generates a public-private key pair. Given a template $\bm{p}$, the client (i) quantizes the features, (ii) encodes the quantized template into a plaintext polynomial, (iii) encrypts the plaintext into a ciphertext using the public key, and (iv) transmits the ciphertext along with metadata of the template to the server. The server then adds the encrypted query to the database. Therefore, the server does not have access to the raw representation of the database at any point. The complete enrollment protocol is described in Algorithm \ref{algo:enrollment} of the Appendix.

\vspace{5pt}
\noindent\textbf{Search Protocol:} Given a query $\bm{q}$, the client (i) quantizes the features, (ii) encodes the quantized query into a plaintext polynomial, (iii) encrypts the plaintext into a ciphertext using the private key, and (iv) transmits the ciphertext to the server. The server computes the encrypted scores as described above and sends them back to the client. The client now decrypts the encrypted scores using the private-key and obtains the index of the nearest match. This index can then be transmitted back to the server depending on the downstream tasks. Therefore, the server does not have access to the raw representation of the query or the matching scores at any point. The complete search protocol is described in Algorithm \ref{algo:search} of the Appendix.

\vspace{5pt}
\noindent\textbf{Computational Complexity} The key technical barrier to realizing homomorphic encryption based representation matching is the computational complexity of the FV scheme, especially ciphertext multiplication. Fundamentally, the addition/multiplication of two integers in the plaintext transforms to addition/multiplication of two polynomials of degree $n$, leading to a $n$-fold and $\mathcal{O}(n^2)$-fold increase\footnote{In practice, can be reduced to $\mathcal{O}(n\log n)$ through number theoretic transforms.} in computational complexity for addition and multiplication, respectively. Therefore, mitigating the number of ciphertext multiplications can lead to large gains in practical utility.

Table~\ref{table:compute} compares the computational complexity of different encoding schemes for secure distance computation. A na{\"i}ve implementation of the FV scheme, i.e., no SIMD, encrypts each element of the representation and performs score matching. Such a scheme has a large computational burden bordering on being impractical for real-world applications. The 1:1 matching scheme in \cite{boddeti2018secure} is specifically designed for vector-vector inner products by encoding an entire $d$-dimensional feature vector into a polynomial. A major computational bottleneck in their scheme is the need for expensive ciphertext \emph{rotations} in order to compute the inner product without access to the individual dimensions of the ciphertext vector. Therefore, this approach scales linearly with the size of the database $m$. In contrast, we observe that our proposed encoding scheme enables \ourmethod~to scale to larger databases due to slower rate of increase in computational complexity by a factor of $\sim \mathcal{O}\left(\frac{d}{n}\right)$.

\subsection{Dimensionality Reduction for Faster FHE Matching}
From the previous subsection we observe that a smaller feature dimension $d$ provides a greater computational savings for $1:m$ matching in comparison to existing approaches. Therefore, to further ease the computational burden of \ourmethod{}, we reduce the dimensionality of a given representation as much as possible, while still retaining accuracy (\textit{i.e.} we attempt to map a representation from the ambient space to its intrinsic dimensionality\footnote{Intrinsic dimensionality~\cite{gong2019intrinsic} is the minimum number of dimensions needed to maintain the information present in the ambient representation.}). For example, we show a reduction of a DeepPrint~\cite{deepprint} fingerprint representation from its 192-dim ambient space to a 32-dim space while losing only 0.1\% rank-1 search accuracy, but getting a $\bm{6\times}$ search speed up within the encrypted domain.

There is growing evidence~\cite{gong2019intrinsic,ansuini2019intrinsic} that representations learned by deep convolutional neural networks are highly redundant, i.e., they lie on a low-dimensional manifold whose intrinsic dimensionality is $20\times$ to $30\times$ smaller than the ambient space that the representation is embedded in. Gong \etal\cite{gong2019intrinsic} further learned a non-liner mapping, dubbed DeepMDS, to compress the representation from the ambient space to its intrinsic space.

Although DeepMDS~\cite{gong2019intrinsic} is able to learn demonstrably better mappings from the ambient to intrinsic space than existing dimensionality reduction schemes, it suffers from two major limitations which limits its direct utility in \ourmethod{}. First, DeepMDS is limited in its generalization capability since it needs to be trained separately for each dataset. Furthermore, it was trained and evaluated on the same dataset. However, real-world application of \ourmethod{} would necessitate a representation compression approach that generalizes across different datasets. Second, DeepMDS optimizes the average pair-wise loss of the samples in a mini-batch without explicitly accounting for the gross imbalance between the number of impostor and genuine pairs. Consequently, the mappings learned by DeepMDS do not focus sufficiently on preserving the local structure of the representation. Both of these drawbacks manifest themselves downstream in the form of loss in feature matching accuracy in the learned intrinsic space, especially when generalizing across different datasets. Therefore, DeepMDS does not serve our goal of compressing the representation while maintaining feature matching performance.

To mitigate the aforementioned limitations we propose \textbf{DeepMDS++}, a deep neural network based approach for compressing the representation. It is comprised of multiple non-linear layers, and is trained to map an embedding from its ambient space to the intrinsic space, such that pairwise distances are preserved. More formally, let $\mathbf{q}\in\mathbb{R}^d$ be a high-dimensional representation in the ambient space and $f(\cdot;\bm{w})$ be the DeepMDS non-linear mapping function with parameters $\bm{w}$. Then, a representation $\mathbf{y}$ in the estimated intrinsic space is computed in accordance with $\mathbf{y} = f(\mathbf{x};\bm{w})$

Let $\bm{G} \in \mathbb{R}^{b_1 \times d}$ and $\bm{I} \in \mathbb{R}^{b_2 \times d}$ be a set of genuine and imposter pair of templates in the ambient space for a mini-batch of size $b=b_1 + b_2$. Similarly, let $\bm{\hat{G}} \in \mathbb{R}^{b_1 \times s}$ and $\bm{\hat{I}} \in \mathbb{R}^{b_2 \times s}$ be the corresponding features in the intrinsic space. Given the pairwise similarities between genuine pairs $\bm{D}_G$ and imposter pairs $\bm{D}_I$ in the ambient space and in the intrinsic space ($\bm{\hat{D}}_G$, $\bm{\hat{D}}_I$), respectively, we optimize the distance between the similarities as follows,
\begin{equation}
    \mathcal{L}_D = \frac{1}{b_1}\|\bm{D}_G - \bm{\hat{D}}_G\|_F^2 + \frac{1}{b_2}\|\bm{D}_I - \bm{\hat{D}}_I\|_F^2
\end{equation}

\vspace{5pt}
\noindent\textbf{Covariance Penalty:} To encourage generalization across datasets, and to explicitly discourage redundancy in DeepMDS++, we adopt a penalty on the covariance of the features. Our loss seeks to minimize the off-diagonal elements of the feature covariance matrix.
\begin{equation}
    \mathcal{L}_c = \|\bm{C} - diag(\bm{C})\|_F^2
\end{equation} where $\bm{C}$ is the covariance matrix computed across our mini-batch of features $[\mathbf{\hat{G}}; \mathbf{\hat{I}}]$ and $diag(\cdot)$ returns a diagonal matrix, with diagonal values of the input matrix.

\vspace{5pt}
\noindent\textbf{Hard Pair Mining:} To allow DeepMDS++ to focus on preserving the local structure of the features in the low-dimensional space, we introduce a hard-mining strategy~\cite{facenet}. This strategy first identifies ``hard pairs" and include them in our mini-batch for training. In particular, we select those pairs for which the pairwise distance is not well preserved after their mapping into the intrinsic space. More formally, the mini-batch indices of the hard genuine pairs $P_G$ and the hard imposter pairs $P_I$ can be computed as:
\begin{equation}
    P_G = argsort(\bm{D}_G - \bm{\hat{D}}_G)
    \quad
    P_I = argsort(\bm{D}_I - \bm{\hat{D}}_I)
\end{equation} where $argsort(\cdot)$ will return the indices of the rows in the features of genuine pairs $\mathbf{G}$ and the impostor pairs $\mathbf{I}$, with the highest error in pairwise distances following their mapping into the intrinsic space (i.e. the hardest pairs).

As we discuss next, experimental results demonstrate that incorporating these solutions aids DeepMDS++ in minimizing the loss of feature matching accuracy when  compressing representations towards their intrinsic dimensionality, ultimately enabling us to perform matching in the encrypted domain more efficiently.


\section{Experiments}
We conduct numerical experiments to evaluate the different components of \ourmethod{}, (i) the proposed encoding scheme that is tailored to $1:m$ matching of features, (ii) DeepMDS++, and (iii) finally the combination the encoding scheme and DeepMDS++.

\subsection{Implementation Details}
DeepMDS++ is implemented in Tensorflow. Details of the network architecture are provided in Appendix~\ref{sec:deepmds}. The network is optimized with Adam~\cite{adam} with an initial learning rate of $3\times10^{-4}$ and with a weight decay of $4\times10^{-5}$. It is trained for 250 epochs using a NVIDIA GeForce RTX 2080 Ti GPU. For hard-pair mining, we used a mini-batch size of $N=4,000$. At the start of training, we mine 50 hard genuine pairs and 50 hard imposter pairs which are augmented with 200 random genuine pairs and 200 random imposter pairs, respectively. Then, we linearly increase the number of hard genuine and hard imposter pairs from 50 to 250 over the 250 training epochs.

The encrypted search in \ourmethod{} is implemented using the SEAL library~\cite{sealcrypto}. For all experiments, we used a 10-core Intel i9-7900X processor running at 3.30 GHz. All evaluations of the search were performed in a single-threaded environment. The three main parameters of the encryption scheme $(n,t,q)$ are set to $n=4,096$, $t=1,032,193$ and $q$ is the default value\footnote{In practice, much smaller values of $q$ are sufficient for our purpose.} in SEAL i.e., a product of 3 very large primes, each 36 bits long.

\begin{figure*}[t]
  \centering
  \begin{subfigure}{0.24\textwidth}
    \centering
    \includegraphics[width=\textwidth]{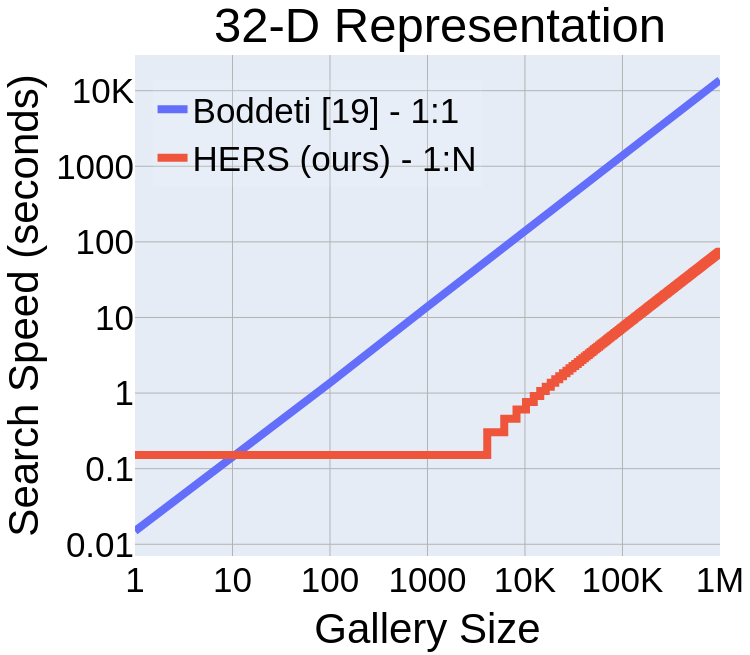}
  \end{subfigure}
  \begin{subfigure}{0.24\textwidth}
    \centering
    \includegraphics[width=\textwidth]{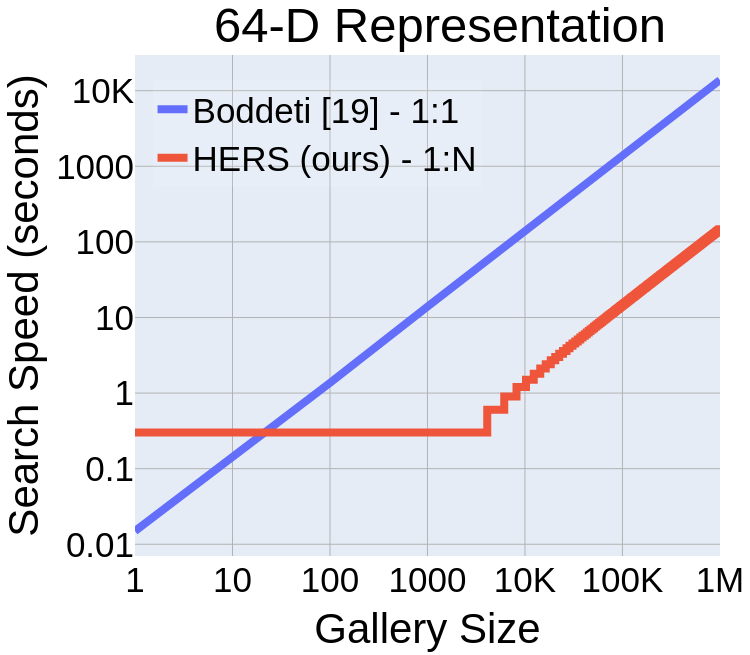}
  \end{subfigure}
  \begin{subfigure}{0.24\textwidth}
    \centering
    \includegraphics[width=\textwidth]{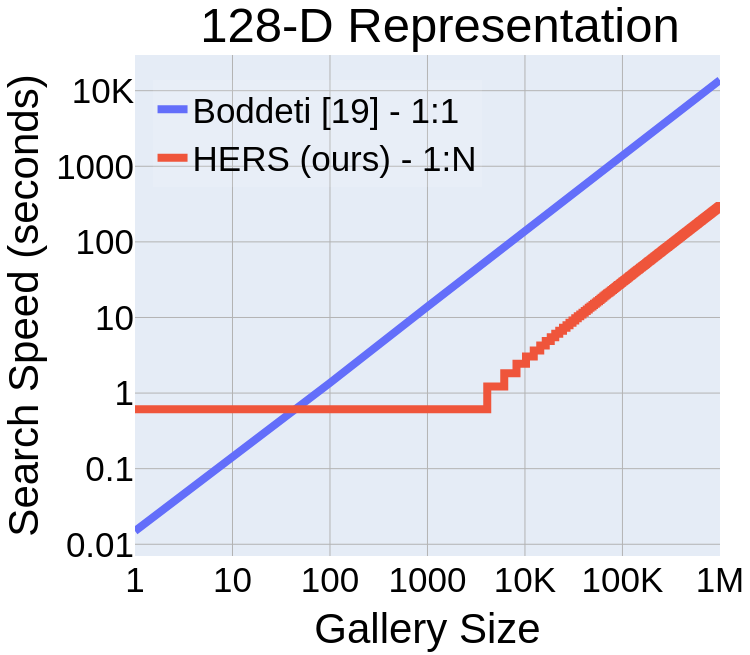}
  \end{subfigure}
  \begin{subfigure}{0.24\textwidth}
    \centering
    \includegraphics[width=\textwidth]{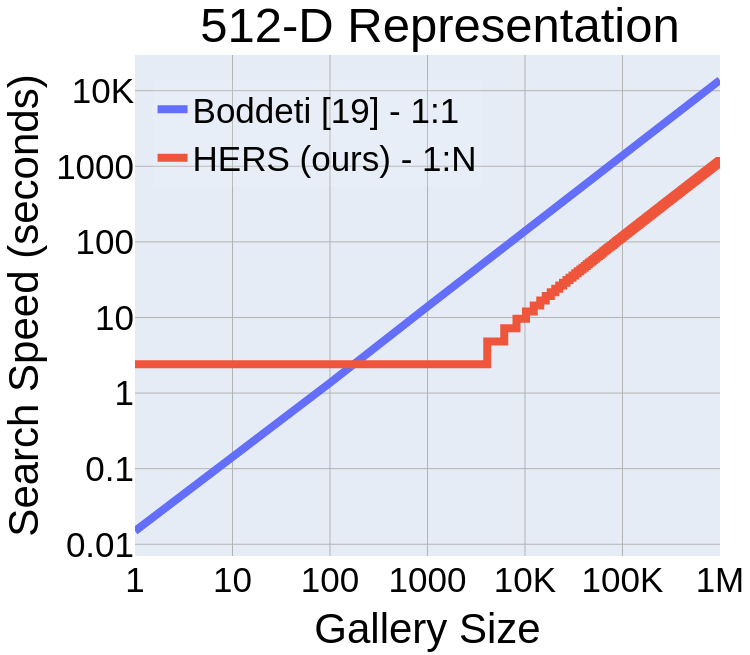}
  \end{subfigure}
  \caption{Computational complexity comparison (log-log scale) of \ourmethod{} with Boddeti~\cite{boddeti2018secure} for $1:m$ matching as a function of gallery size $m$ and different representation dimensionality.
  \label{fig:search_results}}
\end{figure*}

\subsection{Evaluation Datasets}

\begin{itemize}
  \item \textbf{FaceScrub + MegaFace~\cite{megaface}:} MegaFace: 1 million distractor faces; FaceScrub: $\approx 3.5K$ celebrity faces. These are commonly coupled datasets for evaluating face search performance at scale.
  \item \textbf{CASIA~\cite{face1}:} 450K face images from 10K subjects. 100K subset used to train DeepMDS++ prior to its application on MegaFace and FaceScrub, ensuring that our evaluation of DeepMDS++ is cross dataset.
  \item \textbf{NIST SD4~\cite{nist4} + 100 Million Synthetic Background~\cite{mistry2019fingerprint}:} NIST SD4: Contains 2,000 probe/mate inked-rolled fingerprint pairs. 100 Million Synthetic Fingerprints: 100 million synthetic fingerprint database used to evaluate scalability of \ourmethod{}; a separate 100K fingerprints from~\cite{longitudinal} were used to train DeepMDS++.
  \item \textbf{ImageNet ILSVRC 2012:} 1000 classes with 1.28 million training images and 50K validation images. We randomly select 100 classes from the training and validation set for training/testing classification accuracy and precision/recall, and we use the entire validation set for testing precision @ $10$.
\end{itemize}

\subsection{Representation Models}

\begin{itemize}
  \item \textbf{ArcFace~\cite{arcface}:} This model obtained state-of-the-art results on the MegaFace Challenge via (i) architectural refinements, (ii) a well curated training dataset, and (iii) an additive arc margin loss. We extracted 512-dim embeddings using a publicly available~\footnote{\url{https://github.com/deepinsight/insightface\#pretrained-models}}, pre-trained ArcFace model.
  \item \textbf{DeepPrint~\cite{deepprint}:} This model is one of only several DCNNs for extracting deep representations from fingerprints (192-dim features for DeepPrint). DeepPrint matches the accuracy of state-of-the-art commercial matchers by incorporating fingerprint domain knowledge into the training process.
  \item \textbf{Inception ResNet v2~\cite{inception}:} This model which combines inception modules with residual connections is one of the best models on ImageNet 2012 (Top-1 accuracy $80.3\%$). We use a pre-trained model\footnote{\url{https://keras.io/applications/\#inceptionresnetv2}} to extract ImageNet training and validation features (1,536-dim representations).
\end{itemize}

\begin{table}[t]
\centering
\begin{threeparttable}
\caption{\ourmethod{}: Computational Complexity of Search Against 100 Million Gallery\label{table:search-complexity}}
\begin{tabular}{lcccc}
\toprule
Dimensions & 32 & 64 & 128 & 512 \\
\midrule
Time (seconds)\tnote{1} & 740 & 1,480 & 2,960 & 12,120 \\
Memory (GB) & 140 & 280 & 550 & 2,200 \\
\bottomrule
\end{tabular}
\begin{tablenotes}
\item[1] 138,250 seconds and 9 TB at all dimensions for baseline (Boddeti~\cite{boddeti2018secure}). (Using 10 parallel cores)
\end{tablenotes}
\end{threeparttable}
\end{table}

\subsection{Evaluation Protocol and Experimental Results}

To evaluate the efficacy of \ourmethod{}, (i) we benchmark its efficiency at different gallery sizes and with different representation dimensions against Boddeti~\cite{boddeti2018secure} , and (ii) we benchmark the matching and search accuracy at different representation dimensions against DeepMDS \cite{gong2019intrinsic}. Finally, we (iiii) introduce a two stage encrypted search algorithm which combines \ourmethod{}~together with~\cite{boddeti2018secure} to further improve the encrypted search speed at a scale of 100 million.

\begin{figure*}[t]
\centering
\begin{subfigure}{0.32\textwidth}
    \centering
    \includegraphics[width=\textwidth]{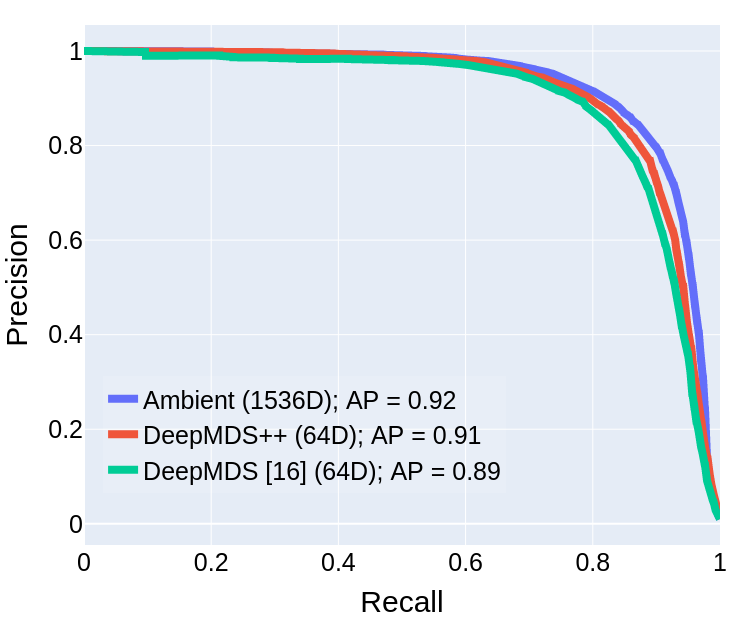}\hfill
\end{subfigure}
\begin{subfigure}{0.32\textwidth}
    \centering
    \includegraphics[width=\textwidth]{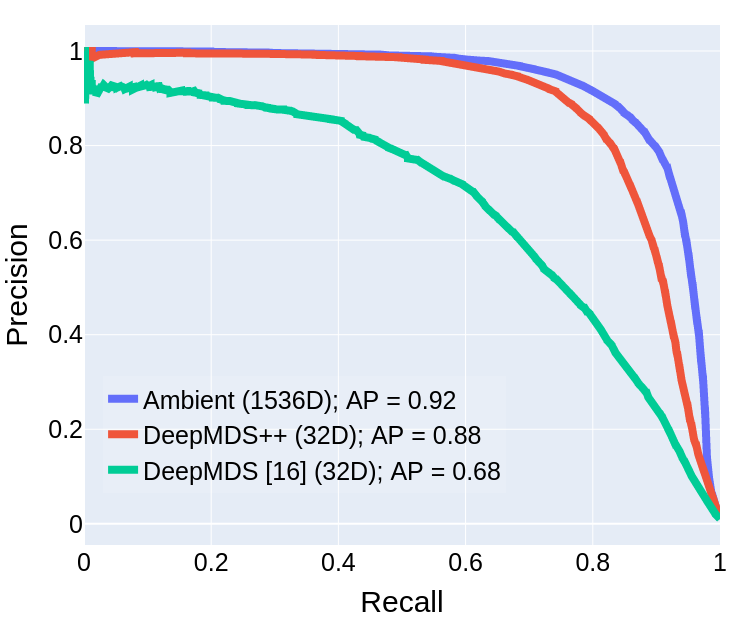}\hfill
\end{subfigure}
\begin{subfigure}{0.32\textwidth}
    \centering
    \includegraphics[width=\textwidth]{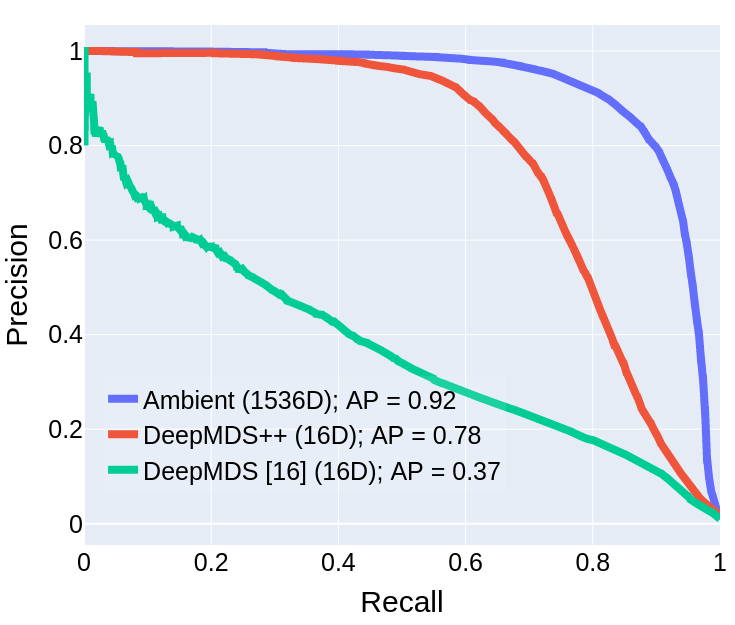}\hfill
\end{subfigure}
\caption{Precision-Recall curves using Inception ResNet V2 on ImageNet 2012\label{fig:pr_curves}}
\end{figure*}

\vspace{3pt}
\noindent\textbf{Efficiency:} Figure~\ref{fig:search_results}, shows the computational complexity of the encryption component of \ourmethod{} with \cite{boddeti2018secure}. Specifically, we consider representations at different dimensions (32-dim, 64-dim, 128-dim, and 512-dim) as we compress them towards their intrinsic dimensionality. This experiment highlights the synergistic benefit of coupling together encryption techniques with dimensionality reduction in order to perform encrypted search at scale. The results indicate that at small gallery sizes 1:1 matching from Boddeti~\cite{boddeti2018secure} is, unsurprisingly more efficient, since it was explicitly designed for 1:1 matching. However, as the gallery increases, \ourmethod{} is more efficient. Admittedly these methods are orders of magnitude slower than matching in the unencrypted domain, but \ourmethod{} is the first practically scalable search over an encrypted database. In terms of memory, for a 64-dim representation, \ourmethod{} requires 5.9 MB for the probe template and 280 GB for a gallery of size 100 million. In comparison~\cite{boddeti2018secure} requires less memory for the probe at 19K, but exhausts 9 TB for the same gallery size. Table~\ref{table:search-complexity} reports the computational complexity of encrypted search, in terms of time and memory, against a gallery of 100 Million for different representation dimensionality. Since HERS is very amenable to parallel processing, we are able to search against 100 million 32-dim representations in 740 seconds ($186\times$ faster than~\cite{boddeti2018secure}), with $64\times$ less memory when using 10 processes (cores) running in parallel. Finally, we note that returning the encrypted scores to the client requires transmitting $m$ integers over the network. Even for a gallery size of 100 million, this only amounts to 100 MB of data. Decrypting the 100 million scores on the client device takes less than 1 second. The inference time for dimensionality reduction is 30 milliseconds. We obtained this inference time on an Intel Core i9-7900X CPU @ 3.30 GHz and 32 GB of RAM.

\vspace{3pt}
\noindent\textbf{Accuracy:} We evaluate the accuracy of DeepMDS++, the dimensionality reduction component of \ourmethod{} in terms of: (i) Rank-1 Face Search Performance on MegaFace (1 million distractors), (ii) Rank-1 Fingerprint Search performance (using NIST SD4 against a gallery of 100 million synthetic fingerprints~\cite{mistry2019fingerprint}), (iii) Top-1 classification accuracy and Precision-Recall curves from a subset of 100 classes of the ImageNet validation set, and (iv) Precision @ 10 (Precision in Top-10 retrieved samples) using the entire ImageNet validation set. For classification experiments on ImageNet, we train a Linear SVM classifier (one-vs-rest) on top of our embeddings. To compute Precision @ 10, we randomly select 10 probes from each validation class, and use the remaining 40 from each class as mates. These mates are combined with the remaining $45,000$ distractors with 900 classes and 50 images / class.

\setlength{\tabcolsep}{7pt}
\begin{table}[h]
    \caption{Face and Fingerprint Search: Rank-1 Accuracy (\%)}
    \centering
    \begin{subtable}{.49\linewidth}
      \centering
      \captionsetup{justification=centering}
      \caption{MegaFace \\(Gallery: 1 Million)\label{table:megaface}}
      \scalebox{0.85}{
      \begin{tabular}{cccc}
        \toprule
        512D & 256D & 128D & 64D \\
        \midrule
        81.4 & 81.4 & 79.0 & 67.9 \\
        \bottomrule
      \end{tabular}}
    \end{subtable}%
    \begin{subtable}{.49\linewidth}
      \centering
      \captionsetup{justification=centering}
      \caption{Fingerprint \\(Gallery: 100 Million)\label{table:msp}}
      \scalebox{0.85}{
      \begin{tabular}{cccc}
        \toprule
        192D & 64D & 32D & 16D \\
        \midrule
        92.3 & 92.3 & 92.2 & 78.6 \\
        \bottomrule
      \end{tabular}}
    \end{subtable}
\end{table}

Table~\ref{table:megaface} and Table~\ref{table:msp} show the Rank-1 accuracy of encrypted face and fingerprint search, respectively, as we compress the representations with DeepMDS++. These results suggest that, practically speaking, we can compress the face representations by a factor of $4\times$ (512-dim to 128-dim) for a performance loss of 2.6\% (81.4\% to 78.8\%). Similarly fingerprint representations can be compressed by a factor of $6\times$ (192-dim to 32-dim) where the performance only drops by $0.1\%$. The 100 Million fingerprint gallery size indicates that DeepMDS++ is incredibly scalable. The degree of compression of the representation by DeepMDS++ closely mirrors the intrinsic dimensionality estimates\footnote{ID estimate code:~\url{https://github.com/human-analysis/intrinsic-dimensionality}} of the respective representations. The intrinsic dimensionalities of ArcFace, DeepPrint and Inception ResNet V2 are 15, 5 and 6 respectively, suggesting that unfolding ArcFace down to 16-dim is about three times as hard as compressing the latter two representations.

\begin{table}[h]
\centering
\captionsetup{justification=centering}
\caption{ImageNet \\ Average Precision (AP) and Top-1 Accuracy (\%)}
    \begin{threeparttable}
	    \begin{tabular}{c|cc||cc}
	        \toprule
	        Dimension\tnote{1,2}\rule{2ex}{0pt} & \multicolumn{2}{c}{DeepMDS~\cite{gong2019intrinsic}} & \multicolumn{2}{c}{DeepMDS++ (ours)} \\
	        \hline
	        \rule{0pt}{2.5ex} & AP & Top-1 (\%) & AP & Top-1 (\%) \\
	        \hline
	        128\rule{0pt}{2ex} & 0.92 & 85.4 & 0.92 & \textbf{85.7} \\
	        64 & 0.89 & 83.4 & \textbf{0.91} & \textbf{85.0} \\
	        32 & 0.68 & 66.1 & \textbf{0.88} & \textbf{81.8} \\
	        16 & 0.37 & 42.7 & \textbf{0.78} & \textbf{71.9} \\
	        \bottomrule
	    \end{tabular}
	    \begin{tablenotes}
            \item[1] Original AP of 0.92 at 1536-dim
            \item[2] Original Top-1 accuracy of 86.2\% at 1536-dim
        \end{tablenotes}
	\end{threeparttable}
	\label{table:precision_classification}
\end{table}

Table~\ref{table:precision_classification} reports the performance of DeepMDS++ and DeepMDS on the ImageNet dataset for the task of image retrieval and image classification, respectively. Figure~\ref{fig:pr_curves} shows the full precision recall curves at dimensions of 1536, 64, 32 and 16 for Inception ResNet v2 embeddings. We make the following observations from these results,

\begin{enumerate}
  \item It is feasible to compress the image representations by large factors for a small performance penalty. For instance, the representation can be compressed by a factor of $24\times$ (1536-dim to 64-dim) for a performance loss of 1\% (92\% to 91\%) in AP and 1\% (86.2\% to 85.0\%) Top-1 accuracy or by a factor of $48\times$ (1536-dim to 32-dim) for a performance loss of 4\% (92\% to 88\%) in AP and 4.5\% (86.2\% to 81.8\%) in Top-1 accuracy.
  \item Representations compressed through DeepMDS++ are able to retain more discriminative information compared to DeepMDS, especially at lower dimensions. For instance, at 32-dim DeepMDS++ obtains an average precision of 88\% compared to 68\% by DeepMDS for image retrieval. Similarly DeepMDS++ Top-1 image classification accuracy is 81.8\% compared to  66.1\% for DeepMDS.
  \item From the precision recall curves, we note that as the Inception ResNet v2 ambient embeddings are compressed to a lower number of dimensions, DeepMDS++ has an increasing advantage over the original DeepMDS~\cite{gong2019intrinsic}.
\end{enumerate}

\begin{figure}[h]
\begin{center}
\includegraphics[width=.9\linewidth]{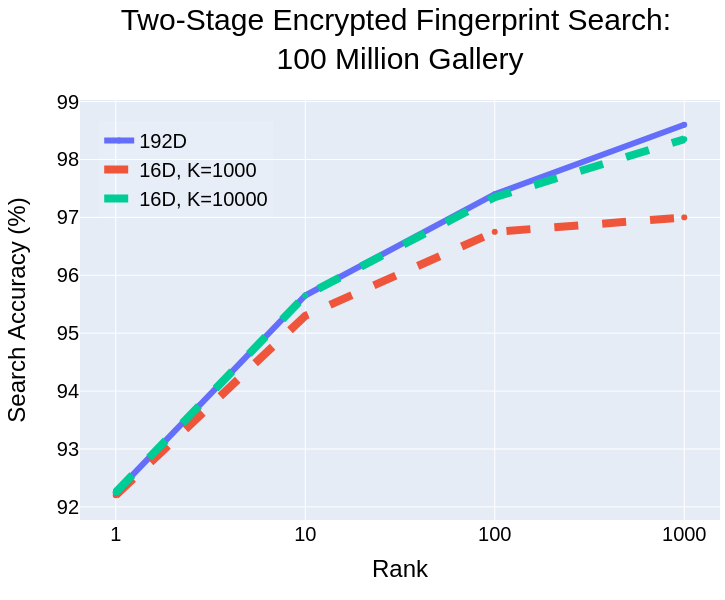}
\caption{DeepPrint rank-level search accuracy on NIST SD4 against a gallery of 100 million using our two-stage encrypted search. When $K=10,000$ in the two-stage search, we can obtain nearly identical search performance as the original full 192-dim DeepPrint representation. \label{fig:two_stage}}
\end{center}
\end{figure}

\vspace{3pt}
\noindent\textbf{Two Stage Search:} In Table~\ref{table:search-complexity}, we demonstrated that \ourmethod{}~obtains encrypted search results against a gallery of 100 million in 740 seconds when the dimensionality of the representation is 32-dim. At 16-dim, this search time could be further reduced to 370 seconds. However, as shown in Table~\ref{table:msp}, the fingerprint search accuracy deteriorates quite heavily when reducing the DeepPrint representation from 32-dim to 16-dim. To leverage the significant time savings from further dimensionality reduction to 16-dim, we propose a two stage encrypted search algorithm. In the first stage, we utilize the 16-dim DeepPrint representation as a ``coarse matcher" in conjunction with \ourmethod{}~to find the top-$K$ candidates. Then, we use the full 192-dim DeepPrint representation in conjunction with the encryption scheme from~\cite{boddeti2018secure} to re-rank the top-$K$ candidates. In this manner, we obtain the benefits of the 16-dim representation for faster search speeds, and by choosing an appropriate value for $K$, we can obtain nearly identical search accuracy to using the 192-dim representation for the full initial search. Note that we use~\cite{boddeti2018secure} for the second stage of the search since it enables matching against the specific indices of the database returned by our first stage search. Figure~\ref{fig:two_stage} shows the rank-level search accuracy for DeepPrint on NIST SD4 against a gallery of 100 million synthetic fingerprints when (i) the initial 192-dim representation is used and (ii) the 16-dim representation is used as a first stage search algorithm, and then the 192-dim representation is used to re-rank. From these results, we observe that when $K=10,000$ for the two-stage search, we can match the accuracy of the full 192-dim representation. This in turn enables cutting the encrypted search time against 100 million fingerprints from 4,500 seconds (at 192-dim) to 500 seconds (370 seconds for stage-1 search, 130 seconds for stage-2 re-ranking) without loss of accuracy. 

Admittedly, one downside of the two stage encrypted search scheme is that it requires encrypting the representations twice (once with the \ourmethod{}~encoding and once with the encoding from~\cite{boddeti2018secure}). This results in the overhead of the 9 TB gallery of~\cite{boddeti2018secure} (for 100 million representations). In general, when designing an encrypted search system, we are looking to optimize along axes of speed, memory, accuracy, and security. Depending on the demands of the particular application, the point at which we land along these axes can be shifted. \ourmethod{}~explores different points along these axes via dimensionality reduction of features (small amounts of accuracy for significant speed savings), and a data-encoding scheme (sacrificing speed and memory at smaller gallery sizes for gains in both at larger galleries). Likewise, the two-stage search schema trades off memory overhead for encrypted search speed and accuracy gains without compromising security.

\vspace{3pt}
\noindent\textbf{Ablation:} Finally, we conduct an ablation on DeepMDS++. Table~\ref{table:ablation} shows the impact of hard-pair mining and the covariance loss on the ability of DeepMDS++ to compress the representations. We observe that hard-pair mining is effective across all dimensions while the covariance loss is more effective around 32 to 64 dimensions where a noticeable benefit is observed. The ability of DeepMDS++ to retain more discriminative information than DeepMDS affords compression to lower dimensions which in turn synergistically aids in improving the efficiency of encrypted search.

\begin{table}[h]
\centering
\caption{Precision @ Rank 10 on ImageNet ILSVRC-2012 Validation Set (\%)}
\label{table:ablation}
\begin{threeparttable}
\begin{tabular}{c|c|c|c}
\toprule
Dim.\tnote{1} & \specialcell{\textbf{DeepMDS++} \\ (proposed)} & \specialcell{w/o \\Hard Mining} & \specialcell{w/o $\mathcal{L}_c$ \\ w/o Hard Mining \\ (DeepMDS \cite{gong2019intrinsic})} \\
\midrule
256 & \textbf{67.6} & 66.1 & 66.1 \\
128 & \textbf{66.5} & 64.1 & 64.3 \\
64 & \textbf{64.0} & 60.1 & 58.9 \\
32 & \textbf{57.6} & 43.6 & 36.4 \\
16 & \textbf{38.0} & 17.0 & 17.1 \\
\bottomrule
\end{tabular}
\begin{tablenotes}
\item[1] {\footnotesize Performance with original 1536-dim Inception ResNet V2 features is 69.7\%.}
\end{tablenotes}
\end{threeparttable}
\end{table}

\section{Security Analysis\label{sec:security}}
We adopt common assumptions in cryptography, i.e., the entities in our system (client and server) are \emph{semi-honest} - each entity ``follows the protocol properly with the exception that it keeps a record of all its intermediate computations"\cite{OdedGoldreich2004}. Under these assumptions, the security of \ourmethod{} is built upon the security of the FV scheme, which in turn is based in the hardness of the \emph{Ring Learning with Errors} problem \cite{lyubashevsky2010ideal}. Practically, this means that the security of our entire protocol hinges upon the fact that the ciphertext cannot be decrypted without access to the private (secret) decryption key which resides only on the client. An attractive property of the FHE scheme \cite{fan2012somewhat} used in this paper is that it offers post-quantum security for an appropriate choice of encryption parameters as outlined in the homomorphic encryption standard \cite{albrecht2019homomorphic}. The encryption parameters $(n,t,q)$ chosen by our experiments correspond to 128-bits of post-quantum security.

The \ourmethod{} system has three sources of vulnerability to attackers, namely, (1) the client device which holds the secret keys, (2) the communication channel between the client and server, and (3) the database which holds the encrypted representations and performs score computation in the encrypted domain. The physical and digital security of the client is most important in order to protect the secret keys. Indeed, if the secret key were obtained from the client, the encrypted templates could be inverted and the templates exposed. Thus the security of the client is of utmost importance. We also note that in HERS, it is possible that some additional protection is rendered to the templates via our dimensionality reduction algorithm (compact feature vectors may be harder to reconstruct useful information from than the original full length representations). The security of the communication channel and the database server are guaranteed by the security of the FV scheme itself.

\vspace{5pt}
\noindent\textbf{Score Inversion:} We now consider an attack based on access to the decrypted scores. We posit that access to the decrypted score without access to the unencrypted features does not allow an attacker to estimate the user's biometric signature. From an adversary's perspective, recovering the unknown feature $\bm{q}\in\mathbb{R}^{d}$ from the scores $\bm{r}=\bm{P}^T\bm{q}$ can be expressed as an optimization problem,
\begin{equation}
    \begin{aligned}
        \min_{\bm{q}} & \quad \|\bm{r} - \bm{P}^T\bm{q}\|_2^2 \\
        s.t. & \quad \bm{q}^T\bm{q} = 1
    \end{aligned}
\end{equation}
\noindent where we assume that the unknown feature is normalized. If the feature is not normalized, the constraint can be removed. The solution of this optimization problem is, $\bm{q}=\left(\bm{P}\bm{P}^T + \lambda\bm{I} \right)^{-1}\bm{P}\bm{r}$. Observe that the solution depends on having access to the raw features $\bm{P}$ in the database. However, without access to the secret decryption keys, the database $\bm{P}$ is available only in the encrypted form. Therefore, even with access to the unencrypted scores, the unknown feature cannot be recovered.

\section{Discussion}
Here we briefly comment on our choice of cryptographic solution, namely, FHE, the limitations induced by our choice and contrast it with other plausible cryptographic solutions. Two other alternative cryptographic solutions that can be employed in lieu of or in conjunction with FHE are:

\vspace{3pt}
\noindent\textbf{Partial Homomorphic Encryption (Paillier Cryptosystem~\cite{paillier1999public}):} This scheme supports only homomorphic additions and is significantly more efficient for scalars than the FV scheme we use. However, it does not support massively vectorized SIMD operations, which is the key source of efficiency in \ourmethod{}.

\vspace{3pt}
\noindent\textbf{Secure Multi-Party Computation~\cite{goldreich1998secure}:} This scheme can be employed to securely compute the nearest neighbors (including matching score and $argmax$ index) by employing multiple parties that communicate secret shares with each other in such a way that no single party can access all the features. This approach trades-off low computation for high communication costs. Furthermore, it requires that the database be split among multiple parties which may not be desirable in some applications. In contrast, the FHE scheme trades-off low-communication costs for high computational costs. The main drawback of the FV scheme is the limited arithmetic operations supported by it, namely addition and multiplication. Therefore, computing non-linear functions like $max$ and $argmax$ are not supported by \ourmethod{}\footnote{Approximate comparison between homomorphically encrypted numbers is now supported by the CKKS FHE scheme~\cite{cheon2020efficient}. In principle, although this capability can be leveraged to compute the \emph{argmax} of the encrypted scores, it is computationally too prohibitive and is a topic of future research.}. In the context of search, as opposed to 1:1 verification, it is often sufficient to protect the gallery and query representations, as opposed to the matching scores as discussed in Section \ref{sec:security}. In such cases, our solution of computing the matching scores in the encrypted domain, having the client decrypt the scores and finally having the server respond with the matched database index should suffice.

\section{Conclusions}
In  this paper, we proposed \ourmethod{}, a scheme for accurate and practical search over homomorphically encrypted representations at scale. The efficiency of \ourmethod{} stems from (i) efficient cryptographic primitives for encrypted matrix-vector products, and (ii) DeepMDS++, a non-linear dimensionality reduction technique to reduce operations in the encrypted domain. The accuracy of \ourmethod{} stems from (i) the exact computations of our cryptographic primitive, and (ii) the effectiveness of DeepMDS++ in maintaining matching performance at large compression factors. Our experimental results demonstrate, for the first time, practical (under 10 minutes) and accurate (within $\approx 2\%$ of unencrypted accuracy) image search (for face, fingerprint and ImageNet) against 100 Million gallery in the encrypted domain.

\section{Acknowledgements}
This research was supported in part by a grant from The NSF Center for Identification Technology Research (CITeR). Vishnu Boddeti was supported in part by award 60NANB18D210 from U.S. Department of Commerce, National Institute of Standards and Technology.

\ifCLASSOPTIONcompsoc
\else
\fi

\ifCLASSOPTIONcaptionsoff
  \newpage
\fi

\bibliography{egbib}
\bibliographystyle{ieeetr}

\newpage

\begin{IEEEbiography}[{\includegraphics[width=1in,height=1.25in,clip,keepaspectratio]{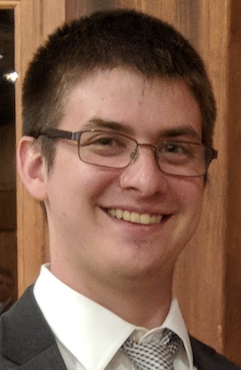}}]{Joshua J. Engelsma} graduated magna cum laude with a B.S. degree in computer science from Grand Valley State University, Allendale, Michigan, in 2016. He obtained a PhD degree in Computer Science and Engineering at Michigan State University, May 2021. His research interests include pattern recognition, computer vision, and image processing with applications in biometrics. He won the best paper award at the 2019 IEEE International Conference on Biometrics (ICB), and the 2020 Michigan State University College of Engineering Fitch Beach Award.
\end{IEEEbiography}

\begin{IEEEbiography}[{\includegraphics[width=1in,height=1.25in,clip,keepaspectratio]{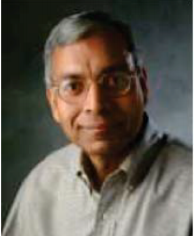}}]{Anil K. Jain} is a University Distinguished Professor in the Department of Computer Science at Michigan State University. He is a Fellow of the ACM, IEEE, IAPR, AAAS and SPIE. His research interests include pattern recognition and biometric authentication. He served as the editor-in-chief of the IEEE Transactions on Pattern Analysis and Machine Intelligence, a member of the United States Defense Science Board and the Forensics Science Standards Board. He has received Fulbright, Guggenheim,  Alexander von Humboldt, and IAPR King Sun Fu awards. He is a member of the United States National Academy of Engineering, a member of The World Academy of Science, and foreign members of the Indian National Academy of Engineering and the Chinese Academy of Sciences.
\end{IEEEbiography}

\begin{IEEEbiography}[{\includegraphics[width=1in,height=1.25in,clip,keepaspectratio]{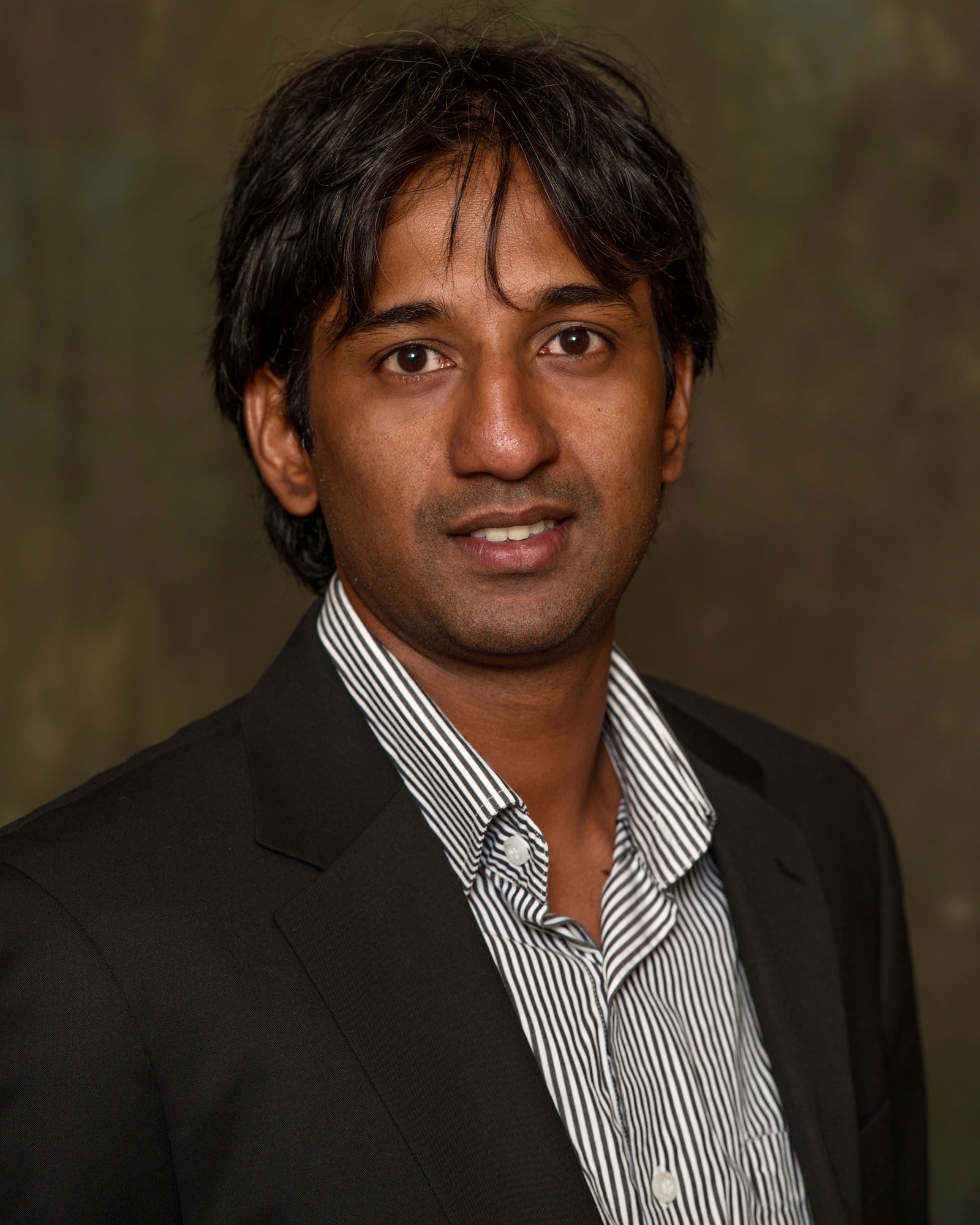}}]{Vishnu Naresh Boddeti} is an Assistant Professor in the computer science department at Michigan State University. He received a Ph.D. degree in Electrical and Computer Engineering program at Carnegie Mellon University in 2013. His research interests are in Computer Vision, Pattern Recognition and Machine Learning. He received the best paper award at BTAS 2013, the best student paper award at ACCV 2018, and the best paper award at GECCO 2019.
\end{IEEEbiography}

\appendices

In this Appendix, we include, (a) details of the base homomorphic encryption scheme \cite{fan2012somewhat} in Section \ref{sec:fv-scheme}, (b) detailed algorithms for the enrollment and search phase of \ourmethod{} in Section \ref{sec:protocols}, (c) security analysis of \ourmethod{} in Section \ref{sec:security}, and (d) architectural details of DeepMDS++ in Section \ref{sec:deepmds}.

\section{Fully Homomorphic Encryption\label{sec:fv-scheme}}
For completeness, we describe the Fan-Vercauteren~\cite{fan2012somewhat} scheme and the associated homomorphic operations, \textit{i.e.}, ciphertext addition and multiplication. These operations will be used in the enrollment and search phase of \ourmethod{}.

\vspace{5pt}
\noindent\textbf{Mathematical Notation: } For $t \in \mathbb{Z}$ a ring $R_t=\mathbb{Z}_t[x]/(x^n+1)$ represents polynomials of degree less than $n$ with the coefficients modulo $t$. The operators $\floor{\cdot}$, $\ceil{\cdot}$ and $\round{\cdot}$ denote rounding down, up and to the nearest integer respectively. The operator $\modulo{\cdot}$ denotes the reduction of an integer by modulo $t$, where the reductions are performed on the symmetric interval $[-t/2,t/2)$. The operators when applied to a polynomial are assumed to act independently on the coefficients of the polynomial. $a\xlongleftarrow{\$}\mathcal{S}$ denotes that $a$ is sampled uniformly from the finite set $\mathcal{S}$. Similarly, $a\xlongleftarrow{}\chi$ denotes that $a$ is sampled from a discrete truncated Gaussian. We note the plaintext polynomial (also called message) as $\bm{m}$ and the ciphertext polynomial as $\bm{ct}$.

\vspace{5pt}
\noindent\textbf{Fan-Vercauteren Scheme \cite{fan2012somewhat}: } The FV scheme encodes integers to polynomials in a ring $R_t$ (see Eq.\ref{eq:query} and Eq.\ref{eq:gallery} for our encoding), referred to as plaintext. Given such a polynomial plaintext, the FV scheme defines the encryption and decryption protocols for such polynomials. The ciphertext is encoded as polynomials in a different ring $R_q$.

Let $\lambda$ be the desired level of security, $w$ the  base to represent numbers in, and $l=\lfloor\log_wq\rfloor$ the number of terms in the decomposition of $q$ into base $w$. Below are the details of the FV scheme in terms of key generation, encryption, decryption, addition and multiplication over encrypted integers.

\begin{tcolorbox}
  \vspace{-0.5cm}
	\begin{algorithm}[H]
	\caption{Key Generation}
    \begin{algorithmic}[1] 
    	\Procedure{GetKeys}{$\lambda$, $l$, $w$, $q$}
			\State Sample: $\bm{\theta}_{sk}\xlongleftarrow{\$}R_2$ \Comment{private (secret) key}
			\State Sample: $\bm{a}\xlongleftarrow{\$}R_q$ and $\bm{e}\xlongleftarrow{}\chi$
			\State $\bm{\theta}_{pk}=([-(\bm{a}\bm{\theta}_{sk}+\bm{e})]_q,\bm{a})$ \Comment{public key}
			\State $\bm{\theta}_{ev} = \emptyset$
			\For{$i=1$ to $l$} \Comment{generate evaluation keys}
				\State Sample: $\bm{a}_i\xlongleftarrow{\$}R_q$, $\bm{e}_i\xlongleftarrow{}\chi$
				\State $\bm{\theta}^i_{ev}=([-(\bm{a}_i\bm{\theta}_{sk}+\bm{e}_i)+w^i\bm{\theta}_{sk}^2]_q,\bm{a}_i)$
				\State $\bm{\theta}_{ev} = \bm{\theta}_{ev} \cup \{\bm{\theta}^i_{ev}\}$
			\EndFor
			\State return $\bm{\theta}_{pk}$, $\bm{\theta}_{sk}$, $\bm{\theta}_{ev}$ \Comment{return all the keys}
		\EndProcedure
		\end{algorithmic}
	\end{algorithm}
\end{tcolorbox}

\begin{tcolorbox}
    \vspace{-0.5cm}
	\begin{algorithm}[H]
	\caption{Encryption}
    \begin{algorithmic}[1] 
    	\Procedure{Encrypt}{$\bm{m}$, $\bm{\theta}_{pk}$, $q$, $t$}
        \State Sample: $\bm{u}\xlongleftarrow{\$}R_2$, $\bm{e}_1\xlongleftarrow{}\chi$ and $\bm{e}_2\xlongleftarrow{}\chi$
        \State $\Delta=\lfloor \frac{q}{t}\rfloor$
        \State $\bm{ct} = ([\Delta\bm{m}+\bm{\theta}_{pk}[0]\bm{u}+\bm{e}_1]_q, [\bm{\theta}_{pk}[1]\bm{u}+\bm{e}_2]_q) = (\bm{ct}[0], \bm{ct}[1])$
        \State return $\bm{ct}$
        \EndProcedure
    \end{algorithmic}
    \end{algorithm}
\end{tcolorbox}

\begin{tcolorbox}
    \vspace{-0.5cm}
	\begin{algorithm}[H]
		\caption{Decryption}
		\begin{algorithmic}[1] 
		\Procedure{Decrypt}{$\bm{ct}$, $\bm{\theta}_{sk}$, $q$, $t$}
		 \State $\bm{pt} = \left[\left\lfloor\frac{t}{q}\left[\bm{ct}[0]+\bm{ct}[1]\bm{\theta}_{sk}\right]_q\right\rceil\right]_t$
		 \State return $\bm{pt}$
	 \EndProcedure
	 \end{algorithmic}
 \end{algorithm}
\end{tcolorbox}

\begin{tcolorbox}
    \vspace{-0.5cm}
	\begin{algorithm}[H]
		\caption{Ciphertext Addition}
		\begin{algorithmic}[1] 
		\Procedure{CipherAdd}{$\bm{ct}_0$, $\bm{ct}_1$, $q$}
		 \State $\bm{ot} = ([\bm{ct}_0[0]+\bm{ct}_1[0]]_q, [\bm{ct}_0[1]+\bm{ct}_1[1]]_q)$
		 \State return $\bm{ot}$
	 \EndProcedure
	 \end{algorithmic}
 \end{algorithm}
\end{tcolorbox}

\begin{tcolorbox}
    \vspace{-0.5cm}
	\begin{algorithm}[H]
		\caption{Ciphertext Multiplication}
		\begin{algorithmic}[1] 
		\Procedure{CipherMultiply}{$\bm{ct}_0$, $\bm{ct}_1$, $w$, $l$, $q$, $t$}
			\State $\bm{c}_0 = \left[\left\lfloor\frac{t}{q}(\bm{ct}_0[0]\bm{ct}_1[0])\right\rceil\right]_q$
		 	\State $\bm{c}_1 = \left[\left\lfloor\frac{t}{q}(\bm{ct}_0[0]\bm{ct}_1[1]+\bm{ct}_0[1]\bm{ct}_1[0])\right\rceil\right]_q$
		 	\State $\bm{c}_2 = \left[\left\lfloor\frac{t}{q}(\bm{ct}_0[1]\bm{ct}_1[1])\right\rceil\right]_q$
		 	\State $\bm{c}_2 = \sum_{i=0}^l c_2^{(i)}w^i$ and compute
		 	\State $\bm{c}_0' = \bm{c}_0 + \sum_{i=1}^l \bm{\theta}_{ev}[i][0]c_2^{(i)}$
		 	\State $\bm{c}_1' = \bm{c}_1 + \sum_{i=1}^l \bm{\theta}_{ev}[i][1]c_2^{(i)}$
		 \State return ($\bm{c}_0', \bm{c}_1'$)
	 \EndProcedure
	 \end{algorithmic}
 \end{algorithm}
\end{tcolorbox}

\section{Protocols\label{sec:protocols}}
Here we describe the detailed algorithms of the two phases in \ourmethod{}, namely, \emph{enrollment} (Algorithm \ref{algo:enrollment}) and \emph{search} (Algorithm \ref{algo:search}). Both of these algorithms are built upon the cryptographic primitives described in Section \ref{sec:fv-scheme}.

\subsection{Enrollment}

Algorithm \ref{algo:enrollment} describes our entire enrollment procedure. The algorithm is designed to handle the scenario where the number of samples in the database $m$ is larger than the ring dimension $n$ (degree of the polynomial). The algorithm also considers a more practical scenario of online enrollment, i.e., we may wish to enroll one gallery feature vector at a time to the encrypted database. For this purpose, we first encrypt an all-zero feature representation and update it with each new gallery we wish to enroll. This is implemented in Lines 12 - 22 of Algorithm~\ref{algo:enrollment} below.

\begin{tcolorbox}
    \vspace{-0.75cm}
    \begin{algorithm}[H]
    \caption{\ourmethod{} Enrollment\label{algo:enrollment}}
    \begin{algorithmic}[1] 
        \State \textbf{Encryption Parameters:} coefficient bit length $b_c$, plaintext modulus $t$, ciphertext modulus $q$, ring dimension $n$
        \State Server initializes empty database $\mathcal{D}_i\leftarrow\emptyset$ $\forall i \in \{1,\dots,d\}$, label set $\mathcal{I}\leftarrow\emptyset$ and database index $k=0$, $v=0$
        \State $\bm{\theta}_{pk}$, $\bm{\theta}_{sk}$, $\bm{\theta}_{ev}$ = $GetKeys(\lambda, l, w, q)$ \Comment{client generates keys}
        \State \textbf{Inputs:} $\bm{id} \in \mathbb{N}^{m}$ and $\bm{q} \in \mathbb{R}^{d\times m}$ \Comment{$m$ feature vectors of dimension $d$ each}
        \For{$i=1$ to $d$}
            \State $v \leftarrow k \mbox{ mod } n$
            \State $\bm{m} \leftarrow BatchEncode(\bm{q}_{i\cdot}; v)$ \Comment{$i$-th dim to $v$-th index of plaintext}
            \State $\bm{ct}_{i} = Encrypt(\bm{m};\bm{\theta}_{pk}, $q$, $t$)$
        \EndFor
        \State $k \leftarrow k + m$ \Comment{increment database index}
        \State Send $(\{\bm{ct}_1,\dots,\bm{ct}_d\}, \bm{id})$ to the server

        \Comment{enrollment at server}
        \State $\mathcal{I} \leftarrow \mathcal{I}\cup\{\bm{id}\}$
        \State $\tilde{v} \leftarrow \left\lceil\frac{k}{n}\right\rceil$
        \If{$\tilde{v} > v$}
        \State $\mathcal{D} \leftarrow \mathcal{D}\cup\{\bm{r}_1,\dots,\bm{r}_d\}$
        \State $v \leftarrow \tilde{v}$
        \State $\bm{r}_i \leftarrow Encrypt(\bm{0};\bm{\theta}_{pk}, $q$, $t$)$ $\forall i \in \{1,\dots,d\}$ \Comment{initialize all zero ciphertext}
        \Else{}
        \For{$i=1$ to $d$} \Comment{enrollment at server}
            \State $\bm{r}_i \leftarrow CipherAdd(\bm{r}_i,\bm{ct}_i; $q$)$
        \EndFor
        \EndIf
    \end{algorithmic}
    \end{algorithm}
\end{tcolorbox}
\begin{tcolorbox}
    \vspace{-0.75cm}
    \begin{algorithm}[H]
    \caption{\ourmethod{} Search\label{algo:search}}
    \begin{algorithmic}[1] 
        \State \textbf{Inputs:} $\bm{q} \in \mathbb{R}^d$, $\mathcal{I}$, $\mathcal{D}$, database index $k$, ring dimension $n$ \Comment{unencrypted query and encrypted database}

        \Comment{authentication at client}
        \For{$i=1$ to $d$}
            \State $\bm{m} \leftarrow BatchEncode(\bm{q}_{i\cdot};0)$ \Comment{$i$-th dim to all indices of plaintext}
            \State $\bm{ct}_{i} = Encrypt(\bm{m};\bm{\theta}_{pk}, $q$, $t$)$
        \EndFor
        \State Send $(\{\bm{ct}_1,\dots,\bm{ct}_d\})$ to the server

        \Comment{authentication at server}
        \State $\mathcal{S}\leftarrow \emptyset$
        \For{$v=1$ to $\left\lceil\frac{k}{n}\right\rceil$}
        \State $\bm{s} \leftarrow Encrypt(\bm{0};\bm{\theta}_{pk}, $q$, $t$)$ \Comment{initialize all zero score ciphertext}
        \For{$i=1$ to $d$}
            \State $\bm{p} \leftarrow CipherMultiply(\bm{ct}_i,\mathcal{D}^v_i; $w$, $l$, $q$, $t$)$
            \State $\bm{s} \leftarrow CipherAdd(\bm{s}, \bm{p}; $q$)$
        \EndFor
        \State $\mathcal{S}\leftarrow\mathcal{S}\cup\{\bm{s}\}$
        \EndFor
        \State Send encrypted scores $\mathcal{S}$ back to client

        \Comment{authentication at client}
        \State $\mathcal{R}\leftarrow \emptyset$
        \For{$l=1$ to $\left\lceil\frac{k}{n}\right\rceil$}
            \State $\bm{r} \leftarrow Decrypt(\mathcal{S}_l;\bm{\theta}_{sk}, $q$, $t$)$
            \State $\mathcal{R}\leftarrow\mathcal{R}\cup\{\bm{r}\}$
        \EndFor
        \State $\mbox{nearest neighbhor} \leftarrow \argmax \mathcal{R}$
    \end{algorithmic}
    \end{algorithm}
\end{tcolorbox}

\subsection{Search}
Algorithm \ref{algo:search} describes our entire search procedure. This includes, query encryption, encrypted score computation, score decryption and argmax on the decrypted scores to find the nearest match.

\section{DeepMDS++\label{sec:deepmds}}
DeepMDS++ is comprised of repeating block units. A single block unit structure is shown in Table~\ref{table:block} and is comprised of two fully-connected layers.
\setlength{\tabcolsep}{10pt}
\begin{table}[!h]
\caption{DeepMDS++ Block Unit ($BU_i$) Architecture\label{table:block}}
\centering
\begin{threeparttable}
\begin{tabular}{c c c}
 \toprule
 Layer Type & Input Dimensions & Output Dimensions\\
\midrule
 \specialcell{Fully Connected \\ (Relu Activation)} & $I_D$\tnote{1} & $I_D$ \\
 \midrule
  \specialcell{Fully Connected \\ (No Activation)} & $I_D$ & $O_D$\tnote{2} \\
 \bottomrule
\end{tabular}
\begin{tablenotes}
\item[1] $I_D$:~Input dimension of current block unit.
\item[2] $O_D$:~Output dimension of current block unit.
\end{tablenotes}
\end{threeparttable}
\end{table}

The number of block units in a given DeepMDS++ model varies, depending on the dimensionality of the original ambient space and the final intrinsic (or as close as possible to intrinsic without losing accuracy) space. The specific repeating block unit structures for the three representation models utilized in this paper are shown (assuming the lowest output dimension reported for each model in the paper) in Table~\ref{table:deepmds_inception} (Inception ResNet v2), Table~\ref{table:deepmds_arcface} (ArcFace), and Table~\ref{table:deepmds_deepprint} (DeepPrint). When we report results for each of these three DeepMDS++ models at higher dimensions in the paper, we simply discard later block units and retrain. In all of our comparisons with the original DeepMDS~\cite{gong2019intrinsic}, we utilize the same repeating block unit architectures for the DeepMDS baseline.

\begin{table}[!h]
\caption{DeepMDS++ (Inception ResNet v2)\label{table:deepmds_inception}}
\centering
\begin{tabular}{c c c}
 \toprule
 Block Unit & \specialcell{Block Input \\ Dimensions ($I_D$)} & \specialcell{Block Output \\ Dimensions ($O_D$)}\\
\midrule
$BU_1$ & 1,536 & 1,024 \\
$BU_2$ & 1,024 & 512 \\
$BU_3$ & 512 & 256 \\
$BU_4$ & 256 & 128 \\
$BU_5$ & 128 & 64 \\
$BU_6$ & 64 & 32 \\
$BU_7$ & 32 & 16 \\
\bottomrule
\end{tabular}
\end{table}

\begin{table}[!h]
\caption{DeepMDS++ (ArcFace)\label{table:deepmds_arcface}}
 \centering
\begin{tabular}{c c c}
 \toprule
 Block Unit & \specialcell{Block Input \\ Dimensions ($I_D$)} & \specialcell{Block Output \\ Dimensions ($O_D$)}\\
\midrule
$BU_1$ & 512 & 256 \\
$BU_2$ & 256 & 128 \\
$BU_3$ & 128 & 64 \\
\bottomrule
\end{tabular}
\end{table}

\begin{table}[!h]
\caption{DeepMDS++ (DeepPrint)\label{table:deepmds_deepprint}}
 \centering
\begin{tabular}{c c c}
 \toprule
 Block Unit & \specialcell{Block Input \\ Dimensions ($I_D$)} & \specialcell{Block Output \\ Dimensions ($O_D$)}\\
\midrule
$BU_1$ & 192 & 128 \\
$BU_2$ & 128 & 64 \\
$BU_3$ & 64 & 32 \\
$BU_4$ & 32 & 16 \\
\bottomrule
\end{tabular}
\end{table}

\end{document}